\documentclass[letterpaper, 10pt, conference]{ieeeconf}  

\usepackage{flushend} 
\usepackage[
  expansion=alltext,
  protrusion=alltext-nott, 
  final 
]{microtype}
\usepackage{amsmath}
\usepackage{amssymb}
\usepackage{siunitx}
\usepackage{graphicx}
\usepackage{cite}  
\usepackage{booktabs}
\usepackage{capt-of} 
\usepackage{placeins} 
\usepackage[nohyperlinks, nolist]{acronym}
\usepackage{bm}
\usepackage{color}
\usepackage[hidelinks]{hyperref}

\graphicspath{{figures}}
\newcommand*{\matr}[1]{\bm{#1}}
\newcommand*{\vect}[1]{\bm{#1}}
\newcommand*{\set}[1]{\mathcal{#1}}
\newcommand*{\Tran}{^{\mathsf{T}}}
\newcommand*{\pinv}{^{\dagger}}
\newcommand*{\R}{\mathbb{R}}

\DeclareMathOperator{\diag}{diag}
\DeclareMathOperator{\timestamp}{time}
\DeclareMathOperator{\obj}{obj}

\acrodef{AR}{augmented reality}
\acrodef{AR/MR}{augmented and mixed reality}
\acrodef{GMM}{Gaussian mixture model}
\acrodef{HRI}{human--robot interaction}
\acrodef{HMD}{head-mounted display}

\IEEEoverridecommandlockouts

\title{\LARGE \bf
  Visual Attention Based Cognitive Human--Robot Collaboration for Pedicle Screw Placement in Robot-Assisted Orthopedic Surgery
}

\author{Chen Chen, Qikai Zou, Yuhang Song, Mingrui Yu, Senqiang Zhu, Shiji Song and Xiang Li
  \thanks{C. Chen, Q. Zou, M. Yu, S. Song and X. Li are with the Department of Automation, Beijing National Research Center for Information Science and Technology, Tsinghua University.
    Y. Song is with the School of Mechanical and Electrical Engineering, Harbin Institute of Technology.
    S. Zhu is with Midea Corporate Research Center and State Key Laboratory of High-end Heavy-load Robots, Midea Group.}%
  \thanks{This work was supported in part by the Science and Technology Innovation 2030-Key Project under Grant 2021ZD0201404, in part by the National Natural Science Foundation of China under Grant U21A20517 and 52075290, in part by Beijing Natural Science Foundation under Grant QY23121, in part by the State Key Laboratory of High-end Heavy-load Robots under Grant HHR2024010426, and in part by the Institute for Guo Qiang, Tsinghua University. Corresponding author: Xiang Li (xiangli@tsinghua.edu.cn)}}

\newcommand{\insertfig}{\includegraphics{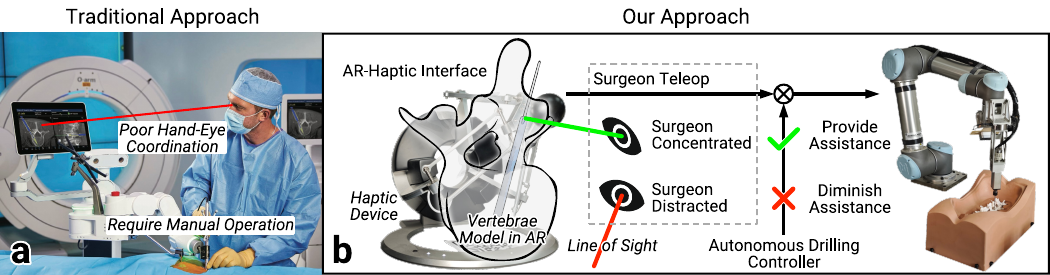}\captionof{figure}{(a) Human--robot interaction in traditional robot-assisted orthopedic surgery. (b) The proposed system, comprising an AR-haptic interface, a gaze-based surgeon attention model, and a shared control scheme. The robot is allowed to carry out tasks with a certain level of autonomy, but it should be allowed to do so when a human is fully concentrated so that the human is always included in the loop.}\label{fig:trailer}}

\makeatletter
\apptocmd{\@maketitle}{\centering\insertfig\setcounter{figure}{1}}{}{}
\makeatother

\begin{document}

\maketitle
\thispagestyle{plain}
\pagestyle{plain}

\begin{abstract}
  Current orthopedic robotic systems largely focus on navigation, aiding surgeons in positioning a guiding tube but still requiring manual drilling and screw placement.
  The automation of this task not only demands high precision and safety due to the intricate physical interactions between the surgical tool and bone but also poses significant risks when executed without adequate human oversight.
  As it involves continuous physical interaction, the robot should collaborate with the surgeon, understand the human intent, and always include the surgeon in the loop.
  To achieve this, this paper proposes a new cognitive human--robot collaboration framework, including the intuitive AR-haptic human--robot interface, the visual-attention-based surgeon model, and the shared interaction control scheme for the robot.
  User studies on a robotic platform for orthopedic surgery are presented to illustrate the performance of the proposed method.
  The results demonstrate that the proposed human--robot collaboration framework outperforms full robot and full human control in terms of safety and ergonomics.
\end{abstract}

\section{Introduction}

The field of robot-assisted orthopedic surgery is witnessing rapid expansion~\cite{dupont21decade,guo22medical}.
Among these, the placement of pedicle screws during spinal fusion stands out as a critical step for ensuring spinal stability~\cite{jiang19new}.
During the procedure, a pilot hole is drilled into the pedicle of the vertebrae, and a screw is tapped into the hole to stabilize the spine.
Existing robotic systems primarily offer navigation capabilities, where the robot positions a guiding tube at the location and orientation of the planned screw entry point.
The surgeon is then required to \emph{manually} drill the hole and place the screw under the robot's guidance.
However, automating bone drilling, which involves complex interactions between the surgical tool and bone, presents a significant challenge due to the high accuracy and safety requirements.

Safety is a main concern of orthopedic robots.
This requires a collaborative approach where the surgeon should be included in the control loop of robot, especially during critical tasks like bone drilling.
While such a surgeon--robot collaboration formulation can guarantee safety and also combine the surgeon's expertise and the robot's precision, it is not trivial to design the collaboration scheme, since the robot needs to understand the surgeon intent then appropriately respond to it exactly.
Too passive reaction manners will shift heavy working loads to the surgeon, and too active ones may result in conflicts between each other.

To address these challenges, this paper proposes a new cognitive human--robot collaboration framework for robot-assisted orthopedic surgery, as shown in Fig.~\ref{fig:trailer}.
The key contributions of this work are summarized as follows:

\begin{itemize}
  \item \textbf{AR-Haptic Interface:}
    A new AR-haptic interface is developed for enhanced surgeon--robot collaboration.
    The \ac{AR} device visualizes the drilling progress and depth of the surgical tool within the bone, while the haptic device enables the surgeon to input the drilling command intuitively by manipulating its end effector, with the perception of drilling force that is aligned with the bone model in the \ac{AR} scene.
    This interface allows for more precise recognition of the surgeon's intent and aids the surgeon to monitor the drilling task from multiple perspectives to make better decisions.

  \item \textbf{Surgeon Attention Model:}
    A surgeon attention model is established to evaluate the surgeon's concentration level.
    The robot will provide assist while the surgeon is fully concentrated to reduce workload.
    The robot will diminish its assistance when the surgeon is not paying enough attention to the task, and the robot will continuously monitor and enforce safety constraints.
    We propose a eye-tracking-based algorithm to recognize the surgeon's attention level.

  \item \textbf{Surgeon--Robot Shared Control:}
    A shared control scheme is proposed to modulate the task allocation between the surgeon and robot based on the intent model. This approach leverages contributions from both sides to enhance the robot-assisted orthopedic surgery.
\end{itemize}

The aforementioned framework offers a cognitive solution for robot-assisted orthopedic surgery, presenting insights into the design of collaborative robotic systems for safety-critical tasks.
It maintains the operator in the loop by monitoring attention levels and amplifies reliable human input when concentration is high, reducing workload.
Experiments and comparative user studies on a collaborative robotic orthopedic platform validate the proposed framework, which outperforms baselines in terms of safety and ergonomics.

\section{Related Works}

\subsection{Human--Robot Interaction in Robotic Orthopedic Surgery}

\subsubsection{Overview}

The majority of orthopedic robotic systems exhibit considerable deficiencies in \ac{HRI}.
Most of these systems have been limited to providing navigational assistance, positioning a guiding tool at a predetermined site to facilitate precise maneuvering of surgical instruments.
This approach requires manual execution of surgical tasks by the surgeon and relies on external monitors for information display, potentially diminishing surgical performance due to its lack of intuitiveness.

Recently, several works have aimed to enhance \ac{HRI} in orthopedic surgical robots.
Futurtec ORTHBOT system, for example, is equipped with an intelligent bone drill that autonomously positions K-wires under the surgeon's supervision of the drilling force~\cite{li21evaluation}.
The Stryker MAKO system can be equipped with a drill or saw, which are held and controlled by the surgeon, allowing for intuitive engagement during surgery~\cite{hagag11rio}.
Lauretti \textit{et al.}\ \cite{lauretti20surgeonrobot} proposed a shared control framework for semi-autonomous pedicle screw fixation, which allows the surgeon to move the robot end effector along the tapping axis using hands-on control and adjust the torque by exerting force on the robot.
Smith \textit{et al.}\ \cite{smith21automated} introduced a robotic system capable of autonomously placing pedicle screws, where the surgeon only needs to oversee the process and intervene when necessary.

\subsubsection{Haptic}

The integration of haptic feedback into orthopedic robotic systems is crucial for providing surgeons with tactile feedback during surgery, enabling precise operation and preventing excessive force application~\cite{taylor16medical}.
Several orthopedic surgical robots have been developed with haptic capabilities.
The Stryker MAKO utilizes haptic feedback to constrain the surgeon's movements according to the interacting forces generated in the virtual haptic environments~\cite{hagag11rio}.
Boschetti \textit{et al.}\ \cite{boschetti05haptic} developed a system for teleoperated spine surgery, where haptic guidance is provided to compensate movements of the vertebra.
Lee \textit{et al.}\ \cite{lee09cooperative} introduced a torque rendering algorithm that provides realistic torque feedback during the tele-drilling of pedicle screws.
Moreover, haptic feedback has started to be incorporated into bilateral telemanipulation systems for minimally invasive surgery~\cite{yilmaz24sensorless}.

\subsubsection{Augmented Reality}

In parallel, \ac{AR}-mediated approaches, as an emerging technology, have been applied to surgical robotics to provide surgeons with intuitive and informative interfaces~\cite{qian20review,suzuki22augmented}.
Iqbal \textit{et al.}\ \cite{iqbal21augmented} developed a system that displays the user interface of an existing orthopedic surgical robot in \ac{AR}, resulting in improved usability and ergonomics.
Tu \textit{et al.}\ \cite{tu22ultrasound} developed a robotic system for cervical pedicle screw placement, where an \ac{AR} surgical scene is constructed and rendered for visualization and navigation.
Schreiter \textit{et al.}\ \cite{schreiter22arsupported} designed an \ac{AR} interface aimed at conditional autonomous robots for pedicle screw placement, enabling surgeons to exert comprehensive oversight and control via an AR interface.

\subsection{Eye-Tracking Based Cognitive Shared Control}

Shared control involves a robot adjusting its level of autonomy based on its understanding of the human's intent and task requirements~\cite{selvaggio21autonomy,losey18review}.
This approach is particularly promising in robot-assisted surgery, where it provides essential assistance while allowing the surgeon to maintain control over the system~\cite{abdelaal20robotics,attanasio21autonomy}.

For effective shared control, the robot must understand the human's intent and then modulate task allocation between the human and the robot accordingly~\cite{javdani18shared,dragan13policyblending}.
Many works have proposed intention inference models to estimate human's intent~\cite{hoffman24inferring}.
However, existing works often overlook the human's cognitive performance during intent assessment, which reflects the concentration level of human.
Wang \textit{et al.}\ \cite{wang22computational} utilized the Yerkes--Dodson law to compute cognitive performance according to the human's utilization ratio, which is the amount of time that the human has been controlling the robot.

Quantitative representation of cognitive performance can be achieved through gaze analysis~\cite{prasov11eye,admoni17social}.
Most existing works on gaze for shared control focus on \emph{intentional} gaze as a control input~\cite{sunny21eyegaze,palinko16robot}.
A series of studies \cite{aronson18eyehand, aronson22gaze, aronson21inferring} have explored the use of \emph{natural} gaze, where hidden-Markov models were applied on gaze signals to predict human intentions in assistive teleoperation tasks.
However, the potential of gaze for cognitive performance assessment in shared control remains underexplored.

\bigskip

In summary, while some progress has been achieved, no existing work has developed a systematic framework that explores the \ac{AR}-Haptic interface and cognitive performance to achieve both the high safety and high autonomy of the drilling task while maintaining an intuitive and efficient collaboration between the surgeon and orthopedic robot.

\section{Methodology}

\subsection{AR-Haptic Human--Robot Interface}

\begin{figure}
  \centering
  \includegraphics{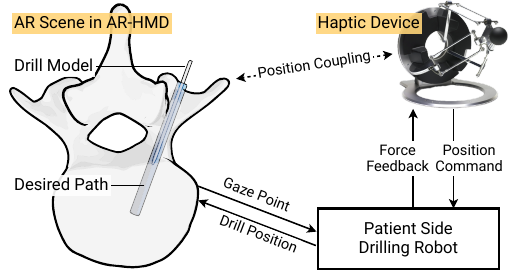}
  \caption{The framework of the proposed human--robot interface. \Ac{AR} scene displays surgical information, while the haptic device enables intuitive operation. Their positions are coupled to enhance hand-eye coordination.} 
  \label{fig:interface}
\end{figure}

First, a new AR-haptic interface is developed for the communication between the surgeon and the robot, as illustrated in Fig.~\ref{fig:interface}.
The haptic device allows the surgeon to intuitively issue drilling commands by manipulating its end effector, which is then converted to the position command of the drill, as detailed in Sec.~\ref{sec:pos}.
The surgeon can also feel the drill's force feedback, enhancing tactile experience.

The \ac{AR} \ac{HMD} is tasked with delivering real-time surgical process information to the surgeon.
To fulfill this purpose, the device renders an AR scene that includes a translucent 3D model of the vertebrae (constructed from CT scans), along with visualizations of the planned pedicle screw trajectory (shown as a transparent blue cylinder) and the real-time drill position (marked by a white drill model).
The eye tracking capability of the \ac{AR} device is utilized to recognize the surgeon's attention level, which is discussed in Sec.~\ref{sec:gaze}.

The position of the rendered \ac{AR} scene is coupled with the haptic device's position by initial registration with a QR code attached to the haptic device.
In this way, the drill tip in the \ac{AR} interface is always aligned with the haptic device's end effector.
Moreover, the vertebrae model is rendered at an enlarged scale, so that the surgeon can have a better view of the details, and can perform more precise movements as the exerted commands is scaled down to match the actual dimensions.
This integrated approach surpasses the capabilities of traditional navigation systems that rely on external displays by providing intuitive and direct visual feedback to the surgeon, enhancing hand-eye coordination.

\subsection{Visual Attention based Human--Robot Collaboration} 

This paper proposes an intent model to specify the cognitive performance of the surgeon during collaborative surgery based on eye-tracking.

\subsubsection{Eye-Tracking Based Human Attention Recognition}
\label{sec:gaze}

\begin{figure}
  \centering
  \includegraphics{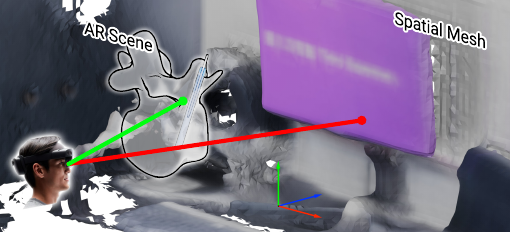}
  \caption{Gaze points are obtained via ray-casting on the \ac{AR} scene or the spatial mesh. The attention level is estimated by the proportion of fixation duration on surgery-related objects within a time window.}
  \label{fig:gaze}
\end{figure}

Eye-tracking data are obtained from the \ac{AR}-\ac{HMD}, consisting of a quaternion representing the gaze orientation within the head coordinate system.
It is subsequently transformed into a direction vector originating from the eye's position within the world coordinate system.
A ray-casting procedure then projects the gaze onto the \ac{AR} scene or the spatial mesh---a mesh representation of the real world, constructed from point clouds captured by the \ac{AR}-\ac{HMD}'s RGB-D camera.
This projection yields the gaze point \(\vect{p}_i \in \R^3\), located either on the AR scene or on a real-world object, as shown in Fig.~\ref{fig:gaze}.

Eye movement data can be categorized into two primary types: saccades and fixations.
In our analysis, we exclusively utilized fixations for attention recognition, filtering out saccadic movements.
The process of fixation segmentation is performed using a similar approach to~\cite{tafaj12bayesian}.
A two-component \ac{GMM} is trained online at regular intervals to model the velocity of the gaze point.
Samples corresponding to the larger component of the \ac{GMM} are identified as saccades, while those associated with the smaller component are classified as fixations.
We define the set of fixations as \(\set{F} = \{\vect{p} \mid \vect{p} \text{ is fixation}\}\).

Moreover, the projection technique enables the semantic annotation of eye gaze through the recognition of the targeted object, denoted by \(\obj(\vect{p})\).
We define a set of objects relevant to surgery, \(\set{S}\), including elements like the vertebrae, drill, and drilling path.
This allows for the formulation of \(\set{A}\), a set of gaze points fixated on surgery-related objects, defined as \(\set{A} = \{\vect{p} \mid \obj(\vect{p})\in \set{S}\}\).

Each gaze point is associated with a specific time stamp, denoted by \(\timestamp(\vect{p})\).
The collection of gaze points within a time window is represented as \(\set{W}(T) = \{\vect{p} \mid \timestamp(\vect{p})\in [t-T, t]\}\), where \(t\) is the current time and \(T\) represents the window's length.

Now, the attention level can be estimated as the proportion of the cumulative duration of fixations on surgical-relevant objects within a time window, which is mathematically expressed as
\begin{equation}
  \alpha(t) = \sum_{\vect{p}_i\in \set{F}\cap \set{A}\cap \set{W}(T)} \frac{\timestamp(\vect{p}_i) - \timestamp(\vect{p}_{i-1})}{T},
\end{equation}
where \(T\) is the time window's length, and \(\timestamp(\vect{p}_i)\) is the time stamp of the gaze point \(\vect{p}_i\).
We further apply an exponential moving average filter to \(\alpha(t)\) to filter out some noise.

\subsubsection{Attention-Driven Human--Robot Collaboration}

A computational human cognitive model is established to dynamically adjust collaboration intensity based on the human's attention level.
This model computes an allocation weight \(w\), which quantifies the degree of robotic assistance at each moment, based on the observed attention of the human.

We propose a \ac{HRI} paradigm particularly for tasks where safety is critical, such as robot-assisted surgery.
The paradigm can be described as
\begin{quote}
  \emph{When the surgeon is not paying enough attention to the task, the robot should diminish its assistance while continuously monitoring and enforcing safety constraints.}
\end{quote}
Thus the allocation shifts from human in control to robot in control while the human's attention level increases.
Such an objective will force the surgeon to focus on the ongoing task and be active in the loop, which is important to ensure safety.

Next, a piecewise linear function is used to implement the aforementioned paradigm, mapping the attention level to the allocation weight by:
\begin{equation}
  w =
  \begin{cases}
    0,                                                   & \text{if } \bar{\alpha} < \alpha_0,                  \\
    \frac{\bar{\alpha} - \alpha_0}{\alpha_1 - \alpha_0}, & \text{if } \alpha_0 \leq \bar{\alpha} \leq \alpha_1, \\
    1,                                                   & \text{if } \bar{\alpha} > \alpha_1,
  \end{cases}
\end{equation}
where \(\bar{\alpha}\) is the filtered attention level, \(\alpha_0\) and \(\alpha_1\) are the thresholds of the attention level, which are determined by the task requirements and the surgeon's cognitive performance. Then, \(w=0\) and \(w=1\) correspond to the manual control and the fully-autonomous mode respectively.

\subsection{Shared Interaction Control}

The constructed weight adjusts the contribution ratio from both sides. Specifically, the robot contributes to the end effector in terms of the drilling depth and speed, and the surgeon contributes through the haptic device. 

\subsubsection{Position Servo of the Patient Side Drilling Robot}
\label{sec:pos}

First, the desired position of the patient side drilling robot \(\vect{x}_{d,\text{ur}}\) is synchronized with the position of the end effector \(\vect{x}\) of the haptic device with a affine transformation, as
\begin{equation}
  \vect{x}'_{d,\text{ur}} = \matr{T}^\text{vertebrae}_\text{robot base} \cdot \diag(1,1,1,k_\text{scale}) \cdot \matr{T}^\text{task space}_\text{vertebrae} \cdot \vect{x}'.
\end{equation}
where \(\vect{x}'_{d,\text{ur}}\) and \(\vect{x}'\) are corresponding homogenous vector of \(\vect{x}_{d,\text{ur}}\) and \(\vect{x}\), \(\matr{T}^\text{vertebrae}_\text{robot base}\) and \(\matr{T}^\text{task space}_\text{vertebrae}\) denote homogenous transform matrices from vertebrae coordinates to robot base frame and from task space to vertebrae coordinates, and \(k_\text{scale}\) is the scaling factor of the vertebrae in \ac{AR} scene.

Such a desired position is achieved with a position control scheme as
\begin{equation}
  \vect{u}_\text{ur} = \matr{K}_{p} \matr{J}_\text{ur}\pinv (\vect{q}) (\vect{x}_\text{ur}-\vect{x}_{d,\text{ur}}), \label{RobotControl}
\end{equation}
where \(\vect{u}_\text{ur}\) is the control input (joint velocities) of the robot, \(\matr{K}_{p}\) is the proportional gain matrix, \(\vect{q}\) are joint angles, and \(\matr{J}_\text{ur}\pinv (\vect{q})\) is the Moore--Penrose pseudoinverse of the Jacobian matrix from joint space to task space.

\begin{figure}
  \centering
  \includegraphics{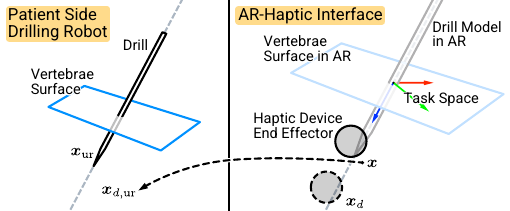}
  \caption{Geometry relationship between the robot and the haptic device.}
  \label{fig:geometry}
\end{figure}

The role of the above controller (\ref{RobotControl}) is to synchronize the robot and the haptic device and, hence, align the contributions on both sides (see Fig.~\ref{fig:geometry}). 

\subsubsection{Shared Interaction Control on Haptic Device}

After the alignment, the shared control can be implemented on the haptic device to enable users to perceive the robot's intentions and assistance tactilely.
The task space is defined as in Cartesian space, where the origin is the start point of the drilling path, and the z-axis points downwards along the drilling axis.

Consider a haptic device with 3 degrees of freedom, its dynamics can be described as
\begin{equation}
  \matr{M}(\vect{\theta})\ddot{\vect{\theta}} + \matr{C}(\vect{\theta}, \dot{\vect{\theta}}) \dot{\vect{\theta}} + \vect{g}(\vect{\theta}) = \vect{\tau}_\text{ext} + \vect{\tau},
\end{equation}
where \(\matr{M}(\vect{\theta})\) is the mass matrix, \(\matr{C}(\vect{\theta}, \dot{\vect{\theta}})\) is the Coriolis matrix, \(\vect{g}(\vect{\theta})\) is the gravitational torques, \(\vect{\tau}_\text{ext}\) denotes external torque and \(\vect{\tau}\) denotes the control input, which is designed to respond to the surgeon's contributions.

To ensure it is safe for the surgeon to inject contributions, a first-order impedance model is formulated as the control objective as
\begin{equation}
  \matr{D} \dot{\vect{x}} + \matr{K} (\vect{x} - \vect{x}_d) = \vect{f}_\text{ext},
\end{equation}
where \(\matr{D}\) and \(\matr{K}\) represent the virtual damping and stiffness matrices to be simulated by the device, \(\vect{x}_d\) is the equilibrium position (i.e., the desired drilling position), and \(\vect{f}_{\text{ext}}\) is the external force applied to the device by the surgeon.

To achieve the objective, the control law on the haptic device is designed as
\begin{equation}
  \vect{\tau} = - \matr{J}\Tran(\vect{\theta}) \bigl(\matr{D} \dot{\vect{x}} + \matr{K} (\vect{x} - \vect{x}_d)\bigr) + \vect{g}(\vect{\theta}).
  \label{simplifiedControl}
\end{equation}
where mass compensation and Coriolis terms are omitted due to the low velocity of the device~\cite{lynch17robot}.
As seen from (\ref{RobotControl}) and (\ref{simplifiedControl}), the proposed formulation requires the surgeon to initiate the task (i.e., \(\vect{x}_d\)), and the robot follows him/her and hence amplifies the surgeon's contributions.
Such a formulation always includes the surgeon in the loop, and the robot provides assistance only when the surgeon is at a high concentration level.

To force the surgeon to re-focus on the task when a distraction arises, control allocation is dynamically adjusted based on the allocation weight \(w\), derived from the human cognitive model, to modulate the desired system behavior: 
\begin{itemize}
  \item For \(w=0\), indicating full human control, the system allows free drag.
  \item For \(w=1\), indicating full automatic control, the device moves downward along the drilling axis at a predetermined constant speed \(v_\text{drill}\).
\end{itemize}

The stiffness matrix \(\matr{K}\) is designed as \(\matr{K}(w) = \diag(k_x, k_y, wk_{z,\text{max}})\).
This configuration ensures that the device maintains high stiffness in movements perpendicular to the drilling axis---represented by \(k_x\) and \(k_y\), with \(k_x\) and \(k_y\) significantly exceeding \(k_{z,\text{max}}\)---thereby constraining the end effector's position within the desired axis.
At \(w=0\), the device permits unrestricted movement along the drilling axis as the stiffness equals zero.

To trace the velocity command, the desired position of the haptic device \(\vect{x}_d\) is updated as
\begin{equation}
  \vect{x}_d = \diag(0,0,1) (v_\text{drill} \matr{K}^{-1}(1)\matr{D}+ \vect{x}).
\end{equation}
In this manner, the haptic device maintains a constant velocity \(v_\text{drill}\) in equilibrium, provided that no external force is exerted upon it, when the robot is in complete control (\(w=1\)).

The feedback force \(\vect{f}_\text{sensor}\) is measured by the force sensor and transformed into the task space coordinates.
To achieve a subtle and smooth transition between human and robot control, the force feedback is scaled down slightly when the robot is automatically controlled, by a function of \(w\): \(\vect{f}_\text{fdbk} = (1 - 0.5w) \vect{f}_\text{sensor}\).

The shared interaction controller of the haptic device integrates these principles:
\begin{equation}
  \vect{u}_\text{haptic} = \matr{J}\Tran(\vect{\theta}) (\matr{K}(w)(\vect{x}_d-\vect{x}) - \matr{D}\dot{\vect{x}} + \vect{f}_\text{fdbk}) + \vect{g}(\vect{\theta}),
\end{equation}
where \(\vect{u}_\text{haptic}\) denotes the driving force of the haptic device.

\section{Experiments}

\begin{figure}
  \centering
  \includegraphics{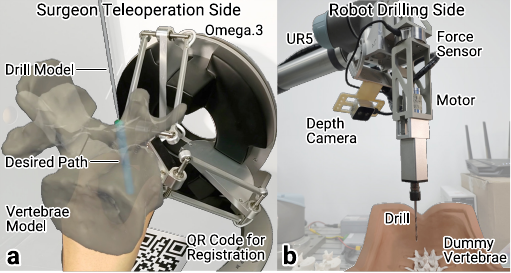}
  \caption{Experimental Setup. (a) The proposed \ac{AR}-haptic interface captured by HoloLens 2. (b) The end-effector design of the robot.}
  \label{fig:setup}
\end{figure}

\begin{figure*}
  \centering
  \includegraphics{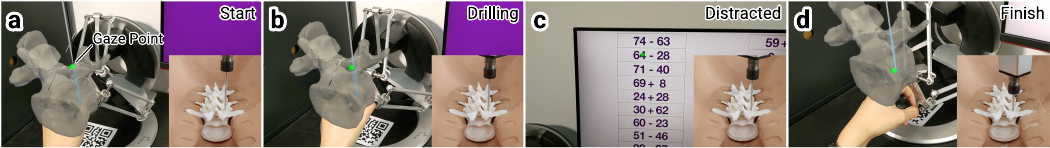}
  \caption{Snapshots from a representative experiment. The participant's gaze point is indicated by a green dot. (a) The participant started drilling. (b) Drilling in progress. (c) Distraction phase: participant shifted attention to solving mental arithmetic problems. (d) Drilling task completed.}
  \label{fig:snapshots}
\end{figure*}

\begin{figure*}
  \centering
  \includegraphics{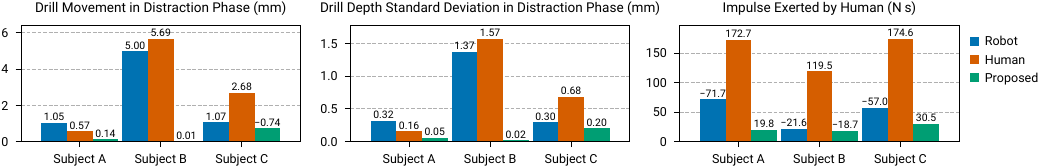}
  \caption{User study results. Three metrics were employed to assess the system's safety and ergonomics, with values closer to zero indicating better performance. During the distraction phase, drill movements are deemed unsafe, thus lower values in the first two metrics indicate enhanced safety. The impulse exerted by the participant throughout the entire task reflects physical load. The proposed shared control method outperformed full robot and full human control in both safety and ergonomics.}
  \label{fig:metrics}
\end{figure*}

\begin{figure}
  \centering
  \includegraphics{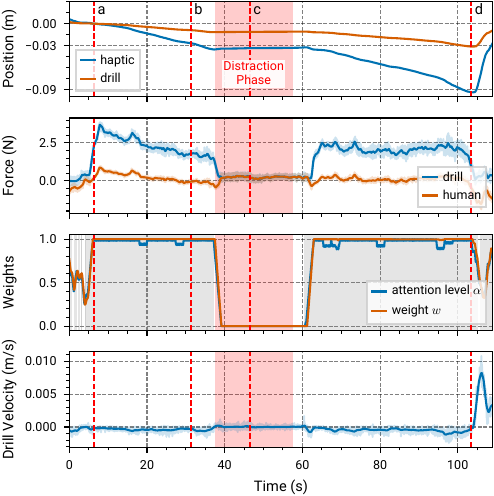}
  \caption{An overview plot of the experiment. The red background represents the distraction phase. In the weights subgraph, the grey background indicates periods when the participant gazed at surgery-related objects. When \(w\approx 1\), the robot was primarily in control, drilling at a predefined speed by applying an additional drilling force; when \(w\approx 0\), the autonomy was transferred to the surgeon, with the robot continuously regulating safety constraints. Time of snapshots in Fig.~\ref{fig:snapshots} are marked as red dotted line.}
  \label{fig:overview}
\end{figure}

In order to evaluate the effectiveness of the proposed framework, experiments were conducted using an orthopedic surgical robot system specifically designed for pedicle screw insertion tasks (see Fig.~\ref{fig:setup}).
The robot featured a UR5 cobot, equipped with a custom bone drill end-effector.
The end-effector is outfitted with an ATI mini40 f/t sensor to monitor drilling forces.
Users interacted with the robot using a Force Dimension Omega.3 haptic device.
The \ac{AR} scene was developed on the Unity engine with MRTK3 and was deployed on Microsoft HoloLens 2 \ac{AR}-\ac{HMD}.
The software system was developed and integrated using the ROS 2~\cite{macenski22robot}, MoveIt 2~\cite{sucanmoveit} and ros2\_control~\cite{chitta17ros_control}.
The interaction controller run at \qty{1}{\kHz} on a PC with Intel i7-12700KF.

\subsection{Experimental Protocol}

A user study was conducted to evaluate the proposed framework.
Participants were asked to drill a \qty{0.03}{\m} deep pilot hole into a synthetic bone model using the proposed robot system.
During the task, a distraction phase is designed to simulate the situation where the surgeon's attention is drawn away from the surgery.
To apply distraction, a series of mental math problems are displayed on an external display in front of the participant.
The participants were asked to concentrate on solving the problems during that phase.
The time of the phase was fixed to \qty{20}{\s}.
The parameters used in the experiments are shown in Table~\ref{tab:params}.

\begin{table}
  \centering
  \caption{Parameters used in the experiments.}
  \label{tab:params}
  \begin{tabular}{cccc}
    \toprule
    Parameter            & Value                     & Parameter          & Value           \\
    \midrule
    \(k_x\), \(k_y\)     & \(\qty{1000}{\N\per\m}\)  & \(T\)              & \(\qty{2}{\s}\) \\
    \(k_{z,\text{max}}\) & \(\qty{50}{\N\per\m}\)    & \(k_\text{scale}\) & \(3\)           \\
    \(\matr{D}\)         & \(\diag(10,10,50)\)       & \(\alpha_0\)       & \(0.1\)         \\
    \(v_\text{drill}\)   & \(\qty{0.001}{\m\per\s}\) & \(\alpha_1\)       & \(0.9\)         \\
    \bottomrule
  \end{tabular}
\end{table}

Instructions were given to the participants before the experiments, during which they were informed of the task and experiment procedure.
The participants were instructed to maintain a high level of attention throughout the task.
After the instruction, the participants were asked to start a trial task to get familiar with the procedure.
Formal experiments were then conducted, where the participants were asked to drill three times, with different human--robot collaboration settings as follows:
\begin{itemize}
  \item Full robot control (\(w=1\)).
  \item Full human control (\(w=0\)).
  \item Proposed visual-attention-based shared control.
\end{itemize}
The sequence of the three settings was randomized to avoid ordering effects, and the participants were not informed of the current experimental setting.

\subsection{Experimental Results}

\subsubsection{Examination of a Representative Experiment}

An in-depth analysis was conducted on a selected experimental result, which is not included in the user study.
This experiment was particularly adjusted to better illustrate the capabilities of the proposed framework, including the explicit display of the gaze point within the \ac{AR} scene.
The snapshots of the whole procedure are shown in Fig.~\ref{fig:snapshots}.

The results of the experiment are shown in Fig.~\ref{fig:overview}.
The raw force and velocity signals are smoothed by a Savitzky--Golay filter~\cite{savitzky64smoothing} with a window size of \qty{1}{\s}.
The time of each snapshot in Fig.~\ref{fig:snapshots} is marked in the plot.
It was seen in the plot that the weight \(w\) gradually increased to 1 when the surgeon was concentrated on the task and decreased to 0 when the surgeon was distracted.

When \(w\approx 1\), the robot was in majority control.
As seen in the force subplot, the automatic controller provided almost all the required drilling force; thus, the force exerted by the surgeon was nearly zero.
The drill moved downward at approximately the desired speed as designed, as shown in the drill velocity subplot.

When \(w\approx 0\), the autonomy was transferred to the surgeon, while the robot constantly regulates safety constraints, such as the movement perpendicular to the drilling axis.
The drilling force and the force exerted by the surgeon were almost identical, as the robot was not providing any assistance.
The drill velocity was dropped to zero, as the surgeon was not actively controlling the drill.

\subsubsection{User Study Results}

The study involved the recruitment of three participants, with the objective of quantitatively evaluating the system’s performance in safety and ergonomic aspects.
Metrics for the assessment included drill movement and its standard deviation of position during the distraction phase, along with the impulse exerted by the surgeon throughout the task.
Safety evaluations were based on metrics of drill movement during the distraction phase, as any movement of the drill while not under the surgeon's supervision is a potential safety risk.
Ergonomics were assessed through the impulse metric derived from the numerical integration of force data over time.
A higher force impulse suggests an increased physical burden and potential for fatigue.

The evaluated metrics of the experiment are shown in Fig.~\ref{fig:metrics}.
Values closer to zero indicating better performance.
Compared to the full robot control and full human control, the proposed shared control method achieved the best performance in safety and ergonomics.



\section{Conclusions}

In this paper, we proposed a cognitive human--robot collaboration framework for robot-assisted orthopedic surgery, based on AR-haptic interface, surgeon visual attention model and shared control.
This work improves the \ac{HRI} experience of orthopedic surgical robots and has the potential to increase surgical efficiency and safety.
Moreover, the proposed framework provides insights into the design of collaborative robotic systems in safety-critical tasks.
Future works will be devoted to the validation of the developed system in clinical trials.





\section*{Acknowledgment}

The authors want to thank Yu Chen for the initial prototype, and thank all subjects for taking part in the human study.


\FloatBarrier
\bibliographystyle{IEEEtran}
\bibliography{ortho}

\end{document}